\documentclass[10pt,twocolumn,letterpaper]{article}

\usepackage{cvpr}              
\usepackage{multirow}

\usepackage{siunitx}
\definecolor{cvprblue}{rgb}{0.21,0.49,0.74}
\definecolor{DeepPink}{rgb}{1,0.078,0.576}
\definecolor{GRAY}{rgb}{0.4, 0.4, 0.4}

\usepackage[pagebackref,breaklinks,colorlinks,allcolors=DeepPink]{hyperref}

\def\paperID{} 
\def\confName{CVPR}
\def\confYear{2026}

\title{\includegraphics[height=1.4em]{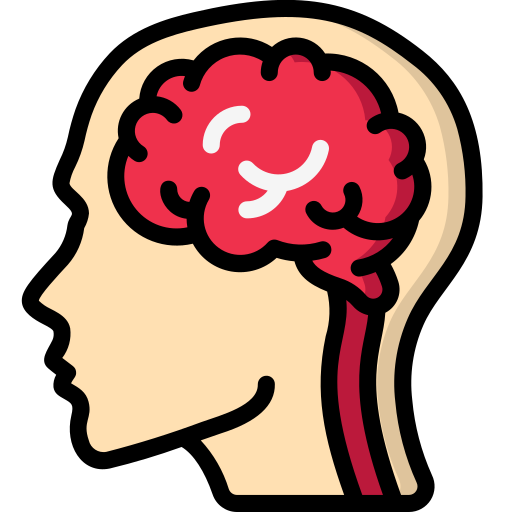}~\textcolor{DeepPink}{LLMind}: Bio-inspired Training-free Adaptive Visual Representations for Vision-Language Models}

\vspace{-20pt}
\author{
Soumyaratna Debnath \quad Bui Duc Manh \quad Zinan Liu \quad Lin Wang$^\dagger$\\
Nanyang Technological University, Singapore\\
{\tt\small soumyara004@e.ntu.edu.sg \quad ducmanh.bui@ntu.edu.sg \quad zinan001@e.ntu.edu.sg \quad linwang@ntu.edu.sg}
}

\begin{document}
\twocolumn[{
\renewcommand\twocolumn[1][]{#1}%
\maketitle
\captionsetup{font=small}
\begin{center}
\vspace{-10pt}
    \centering
    \vspace{-10pt}
    \includegraphics[width=0.95\textwidth]{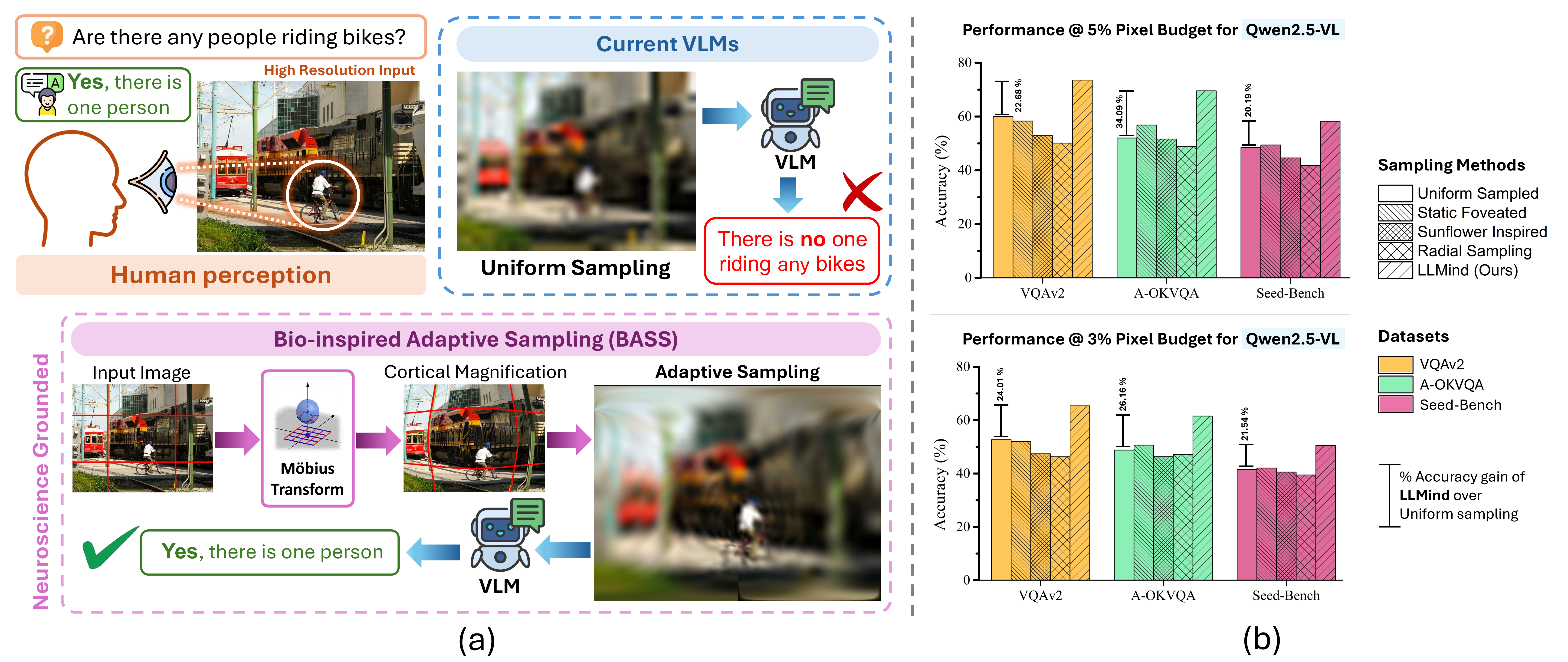}
    \vspace{-10pt}
    \captionof{figure}{\textbf{(a)} Illustration of the underlying principle of our Bio-inspired Adaptive Sampling Strategy (BASS). \textbf{(b)} Performance comparison at 5\% and 3\% pixel budgets using Qwen2.5-VL~\cite{bai2025qwen2} across datasets. Project page: \href{https://empactlab.github.io/LLMind-CVPR-2026/}{https://empactlab.github.io/LLMind-CVPR-2026/}}
\label{fig:teaser}
\end{center}
}] 

\begingroup
\renewcommand\thefootnote{$\dagger$}
\footnotetext{Corresponding author.}
\endgroup

\begin{abstract}
Vision-Language Models (VLMs) typically assume a uniform spatial fidelity across the entire field of view of visual inputs, dedicating equal precision to even the uninformative regions. By contrast, human vision is neither uniform nor static. It is adaptive, selective, and resource-efficient. 
In light of this, we present the \textbf{first} systematic analysis of bio-inspired visual representation methods, providing insights for more efficient and adaptive VLMs. 
We propose \textbf{LLMind} (\textbf{Looking Like the Mind}), a novel \textbf{training-free} framework that mimics foveated encoding and cortical magnification in human vision to achieve \textbf{adaptive}, \textbf{efficient} representations for VLMs under tight pixel budgets. 
Our key idea is to explore a Bio-inspired Adaptive Sampling Strategy (\textbf{BASS}). This empowers us to design a Möbius-parameterized module that \textbf{performs non-uniform sampling while preserving global scene structure.}
On top of BASS, we introduce closed-loop semantic feedback (\textbf{CSF}) via test-time adaptation 
to align the perceptual saliency with the textual information from the frozen VLM. 
We evaluate LLMind against uniform and other sampling baselines across diverse scene-level and region-guided visual question answering (VQA) benchmarks. 
The results show that ours achieves dramatic gains, with average improvements by \textbf{+20}\% on VQAv2, \textbf{+38}\% on Seed-Bench, and \textbf{+37}\% on A-OKVQA compared to uniform sampling under tight pixel budgets. 
More surprisingly, results reveal that LLMind can retain up to \textbf{82}\%, \textbf{92}\% and \textbf{97}\% of the full-resolution performance with only 1\%, 3\% and 5\% of the pixels, respectively. 
Moreover, LLMind is \textbf{lightweight, plug-and-play, and compatible} with existing VLMs without requiring architectural changes. 

\end{abstract}
    
\vspace{-15pt}
\section{Introduction}
\label{sec:intro}

Vision-Language Models (VLMs) have recently demonstrated impressive progress in multimodal reasoning and visual question answering (VQA)~\cite{li2023blip, wang2024qwen2, alayrac2022flamingo, dai2023instructblip}. By jointly encoding visual and textual cues, these models can infer semantic correspondences and generate contextually coherent responses. However, despite these advances, existing models still operate under an assumption:
\textit{perception requires uniform access to every pixel of an image}~\cite{wang2025emulating, schwinn2022behind}.
For instance, typical VLMs, \eg, Qwen~\cite{yang2025qwen3} or LLaVA~\cite{li2024llava} based on attention mechanisms do not employ attention to modulate the spatial sampling of visual inputs~\cite{khan2022transformers}. Attention in transformers redistributes representational emphasis across features, but the underlying image is still uniformly sampled at fixed resolution. Consequently, VLMs allocate equal computational resources to semantically irrelevant background regions and critical foreground details alike \cite{han2022survey, brauwers2021general}. 
While computationally tractable at small scales, these models become inefficient as image resolution increases, forcing models to uniformly downsample and, consequently, discard critical details from all regions of the image.  
Although dynamic tokenization \cite{bolya2022token, rao2021dynamicvit} has recently been introduced to address such issues, it still requires full-resolution input, which limits its suitability for edge systems.

From a neuroscience perspective, uniform downsampling does not reflect how biological vision allocates computational resources~\cite{wang2025emulating, schwinn2022behind}. Instead, the human visual system employs a hierarchical, attention-driven foveation strategy: the fovea provides high-acuity detail in a small attended region, while the peripheral retina conveys coarse, context-rich information that guides subsequent gaze shifts and processing~\cite{strasburger2011peripheral} (Fig.~\ref{fig:teaser} (a)). Through rapid saccadic movements, the visual system dynamically repositions this high-resolution window, sampling the scene adaptively based on task demands~\cite{hayhoe2005eye, land2009looking}. This mechanism reflects the principle of \textbf{cortical magnification}, where perceptually salient regions occupy disproportionately large representational space in the visual cortex~\cite{duncan1984selective}. Perception, therefore, is not a static snapshot but an active optimization process that maximizes information gain while minimizing computational expenditure~\cite{friston2010free}.

Inspired by the foveated encoding and cortical magnification in human vision, an interesting question arises: \textit{\textbf{Can biologically inspired sampling strategies enable VLMs to achieve higher reasoning efficiency and accuracy than conventional uniform sampling under limited pixel budgets?}}
To answer this, we present the first comprehensive analysis of the visual representation strategies in VLMs, evaluating their effectiveness against uniform sampling across multiple VQA benchmarks. Building on this, we propose \textbf{LLMind}, a novel adaptive sampling framework that emulates the human visual system for the visual stimuli.\footnote{Here, visual stimuli correspond to the input image of VLMs.} Our core idea is to explore a Bio-inspired Adaptive Sampling Strategy (\textbf{BASS}) (Sec.~\ref{sec:bass}). BASS serves as a computational analogue of cortical magnification, wherein the foveal region occupies a disproportionately large cortical representation~\cite{daniel1961representation}. To make it possible, we use Möbius transformation~\cite{arnold2008mobius, olsen2010geometry},  a mathematical tool that is parameterized to simulate cortical magnification. As a conformal mapping on the spherical plane,  Möbius transformations perform smooth mappings of the image space (\ie rotation, translation, scaling, inversion) while preserving local geometry. As such, we can dynamically remap the spatial domain of an image, magnifying task-relevant regions with higher sampling density while compressing peripheral areas. 

We optimize the BASS parameters in a training-free loop using a perceptual loss, guided by the semantic output of the frozen VLM. Since the frozen VLM model is non-differentiable, we then introduce closed-loop semantic feedback (\textbf{CSF}) via test-time adaptation using Simultaneous Perturbation Stochastic Approximation~\cite{spall2002multivariate} to estimate gradients directly from input-output interactions (Sec.~\ref{sec:csl}). Unlike prior training-free methods that require access to internal model parameters or gradients of open (\textit{white-box}) VLM \cite{wu2024controlmllm}, our gradient-free CSF strategy can also seamlessly integrate with proprietary (\textit{black-box}) APIs, requiring no architectural modifications or parameter access.

We extensively evaluate LLMind against uniform and other sampling baselines across diverse benchmark datasets, considering two complementary VQA paradigms: \textit{Scene-level VQA} for holistic reasoning and \textit{Region-guided VQA} for localized understanding. LLMind improves the average performance\footnote{Average across 1\%, 3\% and 5\% pixel budgets.} by \textbf{+20\%} on \textit{VQAv2}~\cite{goyal2017making}, \textbf{+38\%} on \textit{Seed-Bench}~\cite{li2023seed}, and \textbf{+37\%} on \textit{A-OKVQA}~\cite{schwenk2022okvqa} compared to uniform sampling under the same pixel budget for scene-level VQA tasks. For region-guided VQA, LLMind achieves an average gain of \textbf{35\%} over uniform sampling, even surpassing the full-resolution accuracy while using only 1\% of the pixels (See Sec.~\ref{sec:results}).

\noindent To summarise, the major contributions of this paper are:
\begin{itemize}
    \item  Inspired by neuroscience-grounded knowledge, we identify a fundamental problem in VLMs and conduct the first comprehensive analysis of the visual representation strategies in VLMs.
    \item We propose BASS that dynamically reallocates pixel budget to perceptually and semantically salient regions as a computational analogue of human vision (Sec.~\ref{sec:bass}).
    \item We introduce the CSF mechanism that aligns visual perception with task-driven reasoning in a training-free, test-time optimization compatible with both white-box and black-box VLMs (Sec.~\ref{sec:csl}).
    \item We validate our method on standard VQA benchmarks, demonstrating consistent improvements in reasoning accuracy under constrained pixel budgets for both scene-level and region-guided settings (Sec.~\ref{sec:results}).  
\end{itemize}

\begin{figure}[t!]
    \centering
    \includegraphics[width=0.85\linewidth]{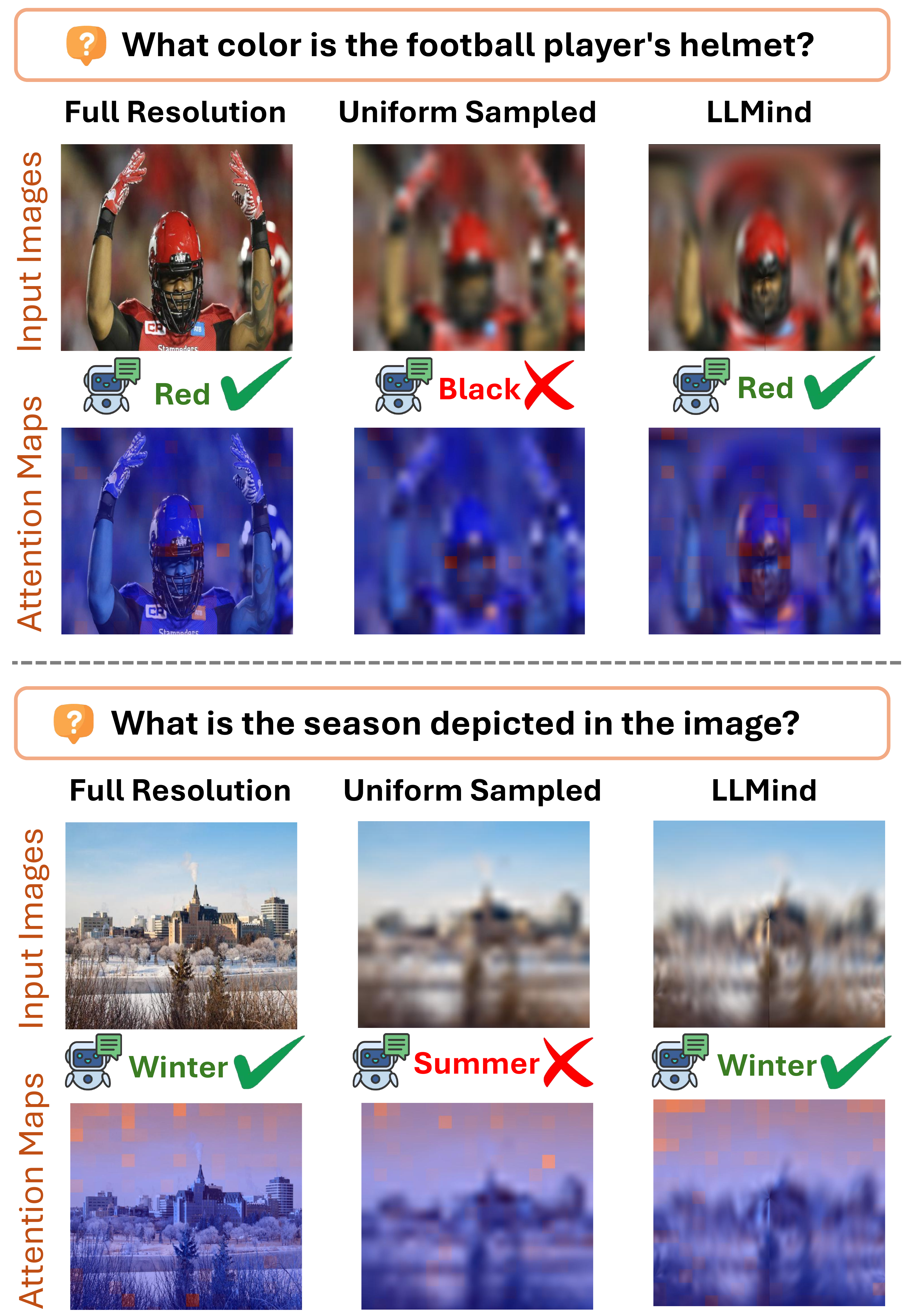}
    \vspace{-5pt}
\caption{\textbf{Qualitative comparison with Qwen2.5-VL on Seed-Bench at 5\% pixel budget.} LLMind adaptively allocates resolution to semantically important regions, preserving visual evidence critical for answering the question. \textit{Zoom in for a better view}.}
    \label{fig:qual_vqa}
    \vspace{-15pt}
\end{figure}
\section{Related Works}
Recently, bio-inspired mechanisms have gained widespread adoption across diverse domains including art, robotics and material design~\cite{manh2025mind, debnath2024scribgen, sakshi2026robotic, debnath2025computational, debnath2023modified}.

\label{sec:related_works}
\noindent \textbf{Multi-Resolution and Non-Uniform Visual Sampling.} Early studies modeled the human visual system to understand the functional limits of peripheral vision \cite{balas2009summary, freeman2011metamers}, but did not translate these insights to AI-based systems. Later research extended this direction by applying foveated texture-based input representations to deep networks, showing that peripheral texture encoding enhances generalization, high-frequency sensitivity, and robustness to occlusion \cite{deza2020emergent, harrington2023coco}. A neuro-computational model further suggested that the advantage of peripheral over central vision arises from the intrinsic utility of peripheral features, which generate a broader representational spreading transform \cite{wang2017central}; however, most of these works were limited to classification tasks~\cite{pramod2022human}. Some works have even explored biologically inspired retinal sampling to mimic foveal vision and reduce visual redundancy for prosthetic or neuromorphic applications~\cite{robert2010biologically}, but these methods remain hardware-bounded, and limited to low-level sensory encoding. Recent studies have examined how multi-resolution processing contributes to robustness and efficiency in vision models \cite{abdulghani2023data, fu2024featup}. In an effort to exploit spatially adaptive representations, several works have explored non-uniform sampling, particularly foveated schemes that densely sample regions near a fixation point while sparsifying the periphery \cite{akbas2017object, lukanov2021biologically, paula2023learning}. However, these approaches often introduced new task-specific architectures rather than utilizing existing pretrained ones. 

\vspace{2pt}
\noindent \textbf{Foveation in Modern Neural Architectures.} Building upon the foundations of non-uniform visual sampling, recent works have integrated foveation principles into modern neural architectures. PerViT \cite{min2022peripheral} modeled peripheral vision by injecting biologically inspired inductive biases into Transformer self-attention, while Killick \etal \cite{killick2023foveation} introduced an end-to-end differentiable foveated vision model using graph convolution and sunflower-patterned sampling. Costa \etal \cite{da2024convolutional} further explored cortical magnification by incorporating a retinal sampling layer that reproduced foveal magnification and radial bias through ganglion-cell-based resampling. However, these approaches remain task-specific, training-dependent, and limited to image classification tasks. Thus, most prior works have focused on simpler single-modality tasks and has limited insight into how biologically inspired sampling can enhance higher-level reasoning in VLMs.

Recently, two works by Gizdov \etal \cite{gizdov2025seeing, gizdov2024variable} demonstrated that variable-resolution sampling improves VQA and detection accuracy in existing Large multimodal models (LMM) under tight pixel budgets. However, their sampling scheme remains static, single-fixation, and does not account for salient or semantic features. In contrast, our framework performs dynamic, inference-time adaptive sampling that accounts for both salient and semantic features under strict pixel constraints, thereby improving performance over Gizdov \etal \cite{gizdov2025seeing, gizdov2024variable}. 

\vspace{2pt}
\noindent \textbf{Training-Free Optimization for Multimodal Adaptation.} Recent works have explored training-free optimization in multimodal systems through visual prompt manipulation~\cite{wu2024controlmllm, jia2022visual}. ControlMLLM~\cite{wu2024controlmllm}, for example,  optimized latent variables over attention maps at test time to guide focus toward region-of-interest cues. Our framework draws inspiration from this paradigm but differs fundamentally: rather than optimizing over attention maps, we optimize the sampling transformation itself, thereby modulating the visual input before encoding. Moreover, while most training-free frameworks, including ControlMLLM, still require access to internal attention maps or gradients, thus operating in a white-box setting. Our method is entirely black-box compatible, functioning seamlessly with closed-source VLMs without any internal model access.

\section{Methodology}
\label{sec:method}

\begin{figure*}[t!]
    \centering
    \includegraphics[width=0.9\textwidth]{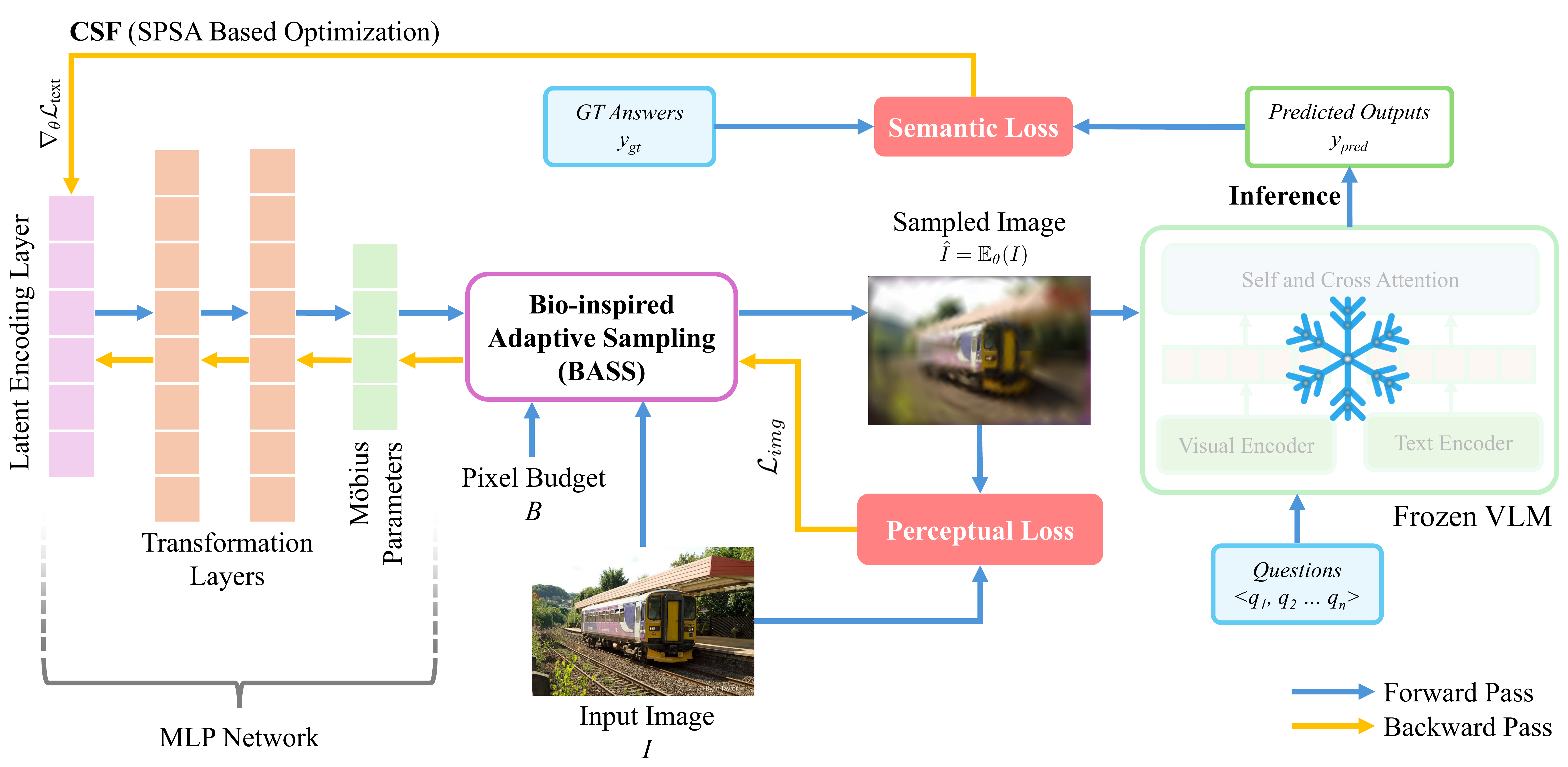}
    \vspace{-5pt}
    \caption{\textbf{Overview of the proposed framework.} Given the input image $I$, the MLP network predicts the Möbius transformation coefficients, which are used by \textit{BASS} (Sec.~\ref{sec:bass}) to produce the sampled image $\hat{I}$. The Perceptual loss $\mathcal{L}_{\text{img}}$ (Eq.~\ref{eq:img_loss}) between $I$ and $\hat{I}$ flows through the network to optimize the MLP parameters. In parallel, a frozen VLM processes a set of questions $q$ to generate predicted answers $y_{\text{pred}}$. These are compared with the ground-truth answers $y_{\text{gt}}$ to obtain the Semantic loss $\mathcal{L}_{\text{text}}$ (Eq.~\ref{eq:text_loss}) for further guiding the optimization of the MLP parameters using SPSA (Eq.~\ref{eq:spsa}).}
    \label{fig:methodology}
    \vspace{-5pt}
\end{figure*}

\noindent \textcolor{black}{\textbf{Overview:}} An overview of the proposed framework is illustrated in Fig.~\ref{fig:methodology}. The core of our design is a \textbf{Bio-inspired Adaptive Sampling Strategy (BASS)} that serves as a front-end sampling module. Based on a Möbius transformation, this generator produces a sampling grid that allocates resolution dynamically, mimicking the foveated structure of the human eye. The transformation parameters are optimized at inference time through a perceptual loss and CSF, analogous to the saccadic eye movement that continuously refines human visual focus. 
Formally, given an image $I \in \mathcal{R}^{H \times W \times 3}$ and a set of questions $q$, the framework produces an adaptively sampled image $\hat{I}$ that maximizes reasoning accuracy under a constrained pixel budget $B$.
{\setlength\abovedisplayskip{2pt}
\setlength\belowdisplayskip{2pt}
\begin{equation}
\hat{I} = \arg \max_{\mathbb{E}_{\theta}(I)} \mathcal{F}(\hat{I}, q, \theta_{f}),
\end{equation}}
where $\mathbb{E}_{\theta}$ denotes the BASS module,parameterized by Möbius transformation coefficients $\theta$ and $\mathcal{F}$ denotes the frozen VLM’s response quality. 

\subsection{Bio-inspired Adaptive Sampling Strategy}
\label{sec:bass}
\begin{figure}[t!]
    \centering
    \includegraphics[width=1.0\linewidth]{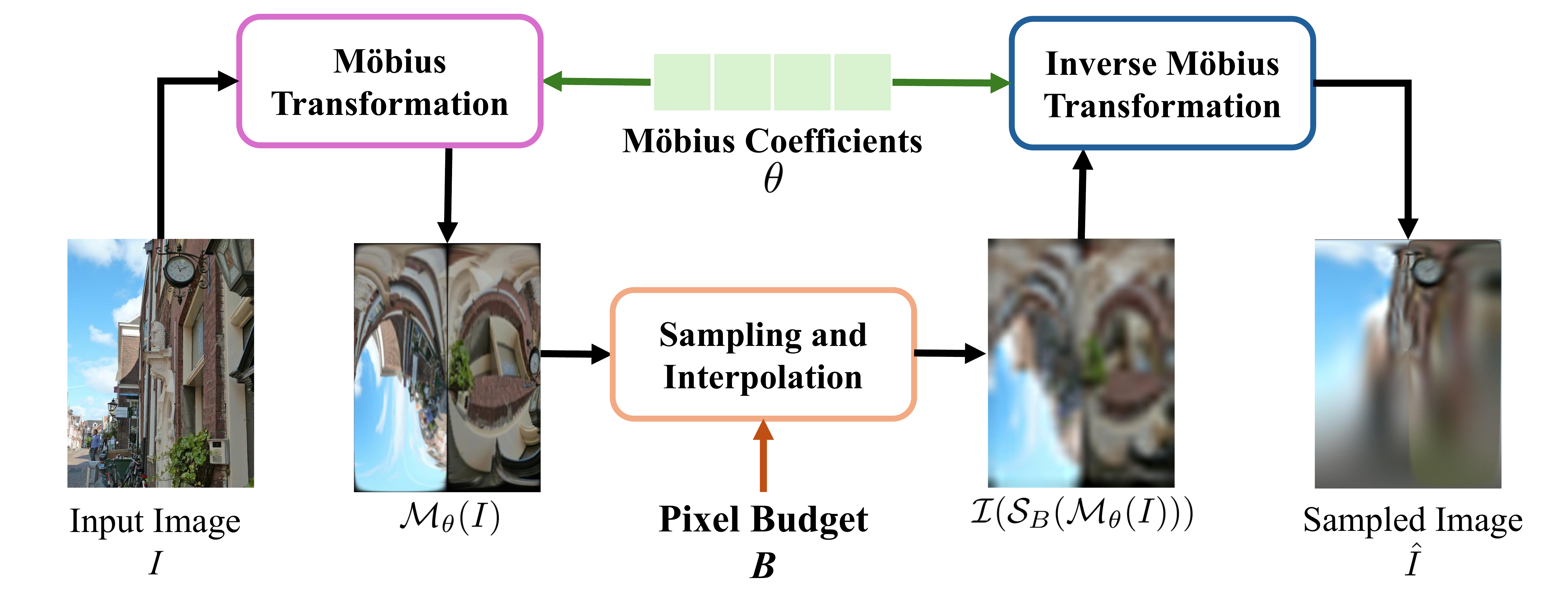}
    \caption{
    \textbf{Illustration of the Bio-inspired Adaptive Sampling Strategy (BASS).} 
    Given an input image $I$, the MLP predicts Möbius parameters $\theta$ to warp the image toward salient regions (Eq.~\ref{eq:mobius}). The warped image is uniformly sampled under pixel budget $B$ through $\mathcal{S}_B(\cdot)$, and then reconstructed to its original resolution via an interpolation operator $\mathcal{I}(\cdot)$. Finally, the inverse transformation (Eq.~\ref{eq:inv_mobius}) restores the global spatial structure, yielding the adaptively sampled image $\hat{I}$.}
    \label{fig:bass}
    \vspace{-10pt}
\end{figure}

\noindent \textcolor{black}{\textbf{Insight.}} Inspired by the non-uniform spatial acuity of the human visual system, we design a Möbius transformation-based Bio-inspired Adaptive Sampling Strategy (BASS) (See Fig.~\ref{fig:bass}). Möbius transformation provides a continuous, conformal mapping of the image plane. It bends the image space so that the cognitive region is magnified while the surrounding areas are smoothly compressed. 

\noindent \textbf{Forward and Inverse Möbius Warp.} We apply a Möbius transformation to the input images via stereographic projection. Given an input image $I$, each pixel corresponds to a spherical direction $\mathbf{s} \in S^2$. We first map $\mathbf{s}$ to the complex plane using north-pole stereographic projection $\Sigma : S^2 \rightarrow \mathbb{C}$, yielding $w = \Sigma(\mathbf{s})$. 
The Möbius transformation used in our model is parameterized by real scalars $a,b,c,d$ (normalized to avoid degenerate matrices), and is applied in the complex domain, and the transformed complex value $z$ is defined as:
{\setlength\abovedisplayskip{2pt}
\setlength\belowdisplayskip{2pt}
\begin{equation}
    z = \frac{a\,w + b}{c\,w + d}
\end{equation}}

\noindent $z$ is then mapped back to a point on the sphere via the inverse stereographic projection $\Sigma^{-1}$, and finally projected back into image coordinates using the standard mapping $\Pi$. Therefore, the overall forward warp $\mathcal{M}(I)$ is:
{\setlength\abovedisplayskip{3pt}
\setlength\belowdisplayskip{3pt}
\begin{equation}
\label{eq:mobius}
    \mathcal{M}(I)[u,v] = I^\star\!\left(\Pi\!\left(\Sigma^{-1}\!\left(
    \frac{d\,\Sigma(\mathbf{s}(u,v)) - b}{-c\,\Sigma(\mathbf{s}(u,v)) + a}
    \right)\right)\right)
\end{equation}}
where $I^\star$ denotes bilinear sampling and $\mathbf{s}(u,v)$ converts the image pixel $(u,v)$ to its corresponding spherical direction.
The inverse warp uses the inverse Möbius $z^{-1}$, defined as:
{\setlength\abovedisplayskip{2pt}
\setlength\belowdisplayskip{2pt}
\begin{equation}
    z^{-1}=\frac{d\,w - b}{-c\,w + a}.,
\end{equation}}
resulting in the inverse transformation $\mathcal{M}^{-1}(I)$:
{\setlength\abovedisplayskip{2pt}
\setlength\belowdisplayskip{2pt}
\begin{equation}
\label{eq:inv_mobius}
\small
    \mathcal{M}^{-1}(I)[u,v] = I^\star\!\left(\Pi\!\left(\Sigma^{-1}\!\left(
    \frac{a\,\Sigma(\mathbf{s}(u,v)) + b}{c\,\Sigma(\mathbf{s}(u,v)) + d}
    \right)\right)\right)
\end{equation}}

\noindent \textbf{Learning the Möbius Parameters.} In our implementation, a lightweight Multilayer Perceptron (MLP) predicts the Möbius parameters, which we represent as a real-valued vector $\theta \in \mathcal{R}^{4}$, corresponding to the real part of the coefficients of the Möbius transformation. 
To incorporate this into an adaptive image sampling framework, we formulate the Sampling Strategy $\mathbb{E}_{\theta}$ as a differentiable mapping:
{\setlength\abovedisplayskip{3pt}
\setlength\belowdisplayskip{3pt}
\begin{equation}
\hat{I} = \mathbb{E}_{\theta}(I) = \mathcal{M}^{-1}_{\theta} \big( \mathcal{I} ( \mathcal{S}_{B} ( \mathcal{M}_{\theta}(I) ) ) \big)
\end{equation}}
where $\mathcal{M}_{\theta}(.)$ and $\mathcal{M}^{-1}_{\theta}(.)$ denote the forward and inverse Möbius transforms parameterized by $\theta$ (Eq.~\ref{eq:mobius} and Eq.~\ref{eq:inv_mobius} respectively), $\mathcal{S}_{B}(.)$ performs uniform sampling at a controlled pixel budget $B$, and $\mathcal{I}(.)$ denotes the interpolation operator (e.g. bilinear interpolation) to restore the image to the original resolution. 

\subsection{Closed-Loop Semantic Feedback}
\label{sec:csl}

\noindent \textcolor{black}{\textbf{Insight.}} We introduce the CSF module to adapt the sampling based on semantics cues via test-time adaptation of the BASS parameters, ensuring that the visual detail is allocated to regions most relevant to the task.
\vspace{-15pt}
\paragraph{Perceptual Loss.} Our perceptual loss is designed to achieve two primary objectives: maintaining global perceptual fidelity and accentuating semantically salient structures. To this end, we define a composite loss function:
{\setlength\abovedisplayskip{3pt}
\setlength\belowdisplayskip{3pt}
\begin{equation}
\label{eq:img_loss}
    \mathcal{L}_{img} = \alpha \cdot \mathcal{L}_{VSI} + \beta \cdot \mathcal{L}_{DISTS} + \gamma \cdot \mathcal{L}_{MSE}
\end{equation}}
\noindent where $\alpha, \beta \text{ and }\gamma$ are weighting coefficients. $\mathcal{L}_{img}$ balances semantic saliency, structural integrity, and optimization stability. It combines a \textbf{Visual Saliency-Induced Index (VSI)}~\cite{zhang2014vsi} term to prioritize human-attentive regions, a \textbf{Deep Image Structure and Texture Similarity (DISTS)}~\cite{ding2020image} metric to enforce perceptual alignment in feature space, and a \textbf{Mean Squared Error (MSE)} regularizer to ensure pixel-level smoothness and stabilize training. 

\vspace{-10pt}
\paragraph{Semantic Loss.} While the perceptual loss $\mathcal{L}_{img}$ preserves visual quality, the challenge is how to improve semantic reasoning accuracy for the given question~\( q \). We therefore integrate textual feedback from the frozen VLM by comparing predicted answers with ground-truth responses.
Given the model's predicted answer~\( y_{pred} \) and reference answer~\( y_{gt} \), we compute the text loss as:
{\setlength\abovedisplayskip{3pt}
\setlength\belowdisplayskip{3pt}
\begin{equation}
\label{eq:text_loss}
\mathcal{L}_{\text{text}} = 1 - \cos\left(E(y_{pred}), E(y_{gt})\right)
\end{equation}}
where \( E(\cdot) \) denotes normalized embeddings obtained from a Sentence Transformer (MiniLM~\cite{wang2020minilm}). This formulation captures semantic similarity rather than exact string matching, making it robust to paraphrasing.
Since the VLM operates as a black box, gradients from \( \mathcal{L}_{\text{text}} \) cannot be directly propagated through the model. To address this, we employ \textbf{Simultaneous Perturbation Stochastic Approximation (SPSA)} to estimate the gradient. Let \( \theta \in \mathbb{R}^4 \) represent the Möbius transformation parameters. The gradient estimate is computed as:
{\setlength\abovedisplayskip{3pt}
\setlength\belowdisplayskip{3pt}
\begin{equation}
\label{eq:spsa}
\nabla_{\theta} \mathcal{L}_{\text{text}} \approx \frac{\mathcal{L}_{\text{text}}(\theta + \delta \Delta) - \mathcal{L}_{\text{text}}(\theta - \delta \Delta)}{2\delta}
\end{equation}}
where \( \delta > 0 \) is a small perturbation size, and \( \Delta \in \mathbb{R}^4 \) is a random perturbation vector with components sampled from a symmetric Bernoulli distribution, i.e., \( \Delta_i \sim \{\pm1\} \) with equal probability.

Furthermore, to improve training efficiency, we employ an adaptive question selection strategy that focuses on challenging examples. Let \( \mathcal{Q} = \{q_1, \dots, q_N\} \) be the set of all questions and \( w_i^{(t)} \) be the weight associated with question \( q_i \) at optimization step \( t \). We initialize weights uniformly: \( w_i^{(0)} = 1/N \). After each evaluation, weights are updated exponentially based on performance:
$w_i^{(t+1)} = w_i^{(t)} \cdot \exp\left(\eta \cdot \frac{\mathbb{I}[q_i \text{ is wrong}]}{N}\right)$,
where \( \eta > 0 \) is a learning rate and \( \mathbb{I}[\cdot] \) is the indicator function. Questions for SPSA gradient estimation are then sampled proportionally to their current weights $P(\text{select } q_i) = \frac{w_i}{\sum_{j=1}^N w_j}.$
We formulate the final optimization objective as a weighted combination of the perceptual and semantic losses. Let \( \mathcal{L}_{\text{img}} \) be the perceptual loss and \( \beta > 0 \) a balancing coefficient. The combined update direction is:
{\setlength\abovedisplayskip{2pt}
\setlength\belowdisplayskip{2pt}
\begin{equation}
g_{\text{total}} = \nabla_{\theta} \mathcal{L}_{\text{img}} + \beta \cdot \frac{\|\nabla_{\theta} \mathcal{L}_{\text{img}}\|}{\|\nabla_{\theta} \mathcal{L}_{\text{text}}\|} \nabla_{\theta} \mathcal{L}_{\text{text}}
\end{equation}}
where gradient norms are computed for numerical stability. This enables gradient-free optimization while maintaining focus on semantically challenging cases, even when the VLM's internal gradients are inaccessible.

\begin{table*}[t!]
\centering
\caption{\textbf{Quantitative comparison of bio-inspired and uniform sampling strategies on scene-level VQA benchmarks under tight-budget constraints (1\%, 3\%, 5\%)}. We report full-resolution accuracy, sampled accuracy at each budget, accuracy gain over uniform sampling, and accuracy retained relative to the full-resolution baseline.}
\vspace{-5pt}
\resizebox{0.95\textwidth}{!}{
\begin{tabular}{ccclccccccccc}
\toprule
\multirow{3}{*}{\textbf{Dataset}} & \multirow{3}{*}{\textbf{Model}} & \multirow{2}{*}{\textbf{Accuracy (\%)}} & \multirow{3}{*}{\textbf{Sampling Method}} &
\multicolumn{3}{c}{\textbf{Accuracy Sampled}} &
\multicolumn{3}{c}{\textbf{$\Delta$ Accuracy Gain (\%)}} &
\multicolumn{3}{c}{\textbf{Accuracy Retained}} \\
& & \multirow{2.5}{*}{\textbf{Full Res.}}& & \multicolumn{3}{c}{\textbf{(\%)}} &
\multicolumn{3}{c}{\textbf{over Uniform Sample}} &
\multicolumn{3}{c}{\textbf{(\% of Full Res.)}}\\
\cmidrule(lr){5-7} \cmidrule(lr){8-10} \cmidrule(lr){11-13}
 & & & & \textbf{1\%} & \textbf{3\%} & \textbf{5\%} & \textbf{1\%} & \textbf{3\%} & \textbf{5\%} & \textbf{1\%} & \textbf{3\%} & \textbf{5\%} \\
\midrule

\multirow{16}{*}{\rotatebox{90}{\textbf{VQAv2}}}
& \multirow{5}{*}{Qwen2.5-VL} & \multirow{5}{*}{86.96}
  & Uniform Sampling                                & 47.31 & 52.71 & 59.94 & -- & -- & -- & 54.40 & 60.61 & 68.92 \\
& & & Static Foveated & 46.18 & 51.96 & 58.30 & -2.38 & -1.42 & -2.73 & 53.10 & 59.75 & 67.04 \\
& & & Sunflower Inspired  & 43.77 & 47.41 & 52.85 & -7.48 & -10.05 & -11.82 & 50.33 & 54.51 & 60.77 \\
& & & Radial Sampling & 44.79 & 46.25 & 50.12 & -5.32 & -12.25 & -16.38 & 51.50 & 53.18 & 57.63 \\
& & & \textbf{LLMind (Ours)} & \textbf{55.05} & \textbf{65.37} & \textbf{73.54} & \textbf{+16.36} & \textbf{+24.01} & \textbf{+22.68} & \textbf{63.31} & \textbf{75.17} & \textbf{84.56} \\
\cmidrule(lr){2-13}
& \multirow{5}{*}{SmolVLM} & \multirow{5}{*}{80.01}
  & Uniform Sampling               & 46.38 & 53.28 & 59.06 & -- & -- & -- & 57.96 & 66.59 & 73.81 \\
& & & Static Foveated   & 44.24 & 51.64 & 58.43 & -4.61 & -3.07 & -1.06 & 55.29 & 64.54 & 73.02 \\
& & & Sunflower Inspired   & 46.45 & 50.55 & 56.38 & +0.15 & -5.12 & -4.53 & 58.05 & 63.17 & 70.46 \\
& & & Radial Sampling & 42.99 & 47.07 & 50.92 & -7.30 & -11.65 & -13.78 & 53.73 & 58.83 & 63.64 \\
& & & \textbf{LLMind (Ours)} & \textbf{58.88} & \textbf{70.40} & \textbf{76.46} & \textbf{+26.95} & \textbf{+32.13} & \textbf{+29.46} & \textbf{73.59} & \textbf{87.98} & \textbf{95.56} \\
\cmidrule(lr){2-13}
& \multirow{5}{*}{LLaVA-OneVision} & \multirow{5}{*}{73.34}
  & Uniform Sampling               & 40.98 & 49.15 & 57.60 & -- & -- & -- & 55.87 & 67.01 & 78.53 \\
& & & Static Foveated   & 42.27 & 42.88 & 45.37 & +3.14 &	-12.75 & -21.23& 57.63 & 58.46 & 61.86 \\
& & & Sunflower Inspired   & 42.29 & 46.02 & 50.51 & +3.19& -6.36&	-12.30&	57.66& 62.74& 68.87
 \\
& & & Radial Sampling & 40.10 & 42.40 & 45.25 & -2.14&	-13.73&	-21.44&	54.67&	57.81&	61.69
 \\
& & & \textbf{LLMind (Ours)} & \textbf{47.00} & \textbf{54.49} & \textbf{62.38} & \textbf{+14.69} & \textbf{+10.86} & \textbf{+8.29} & \textbf{64.08} & \textbf{74.29} & \textbf{85.05} \\
\midrule

\multirow{16}{*}{\rotatebox{90}{\textbf{A-OKVQA}}}
& \multirow{5}{*}{Qwen2.5-VL} & \multirow{5}{*}{85.67}
  & Uniform Sampling                                & 44.31 & 48.77 & 52.01 & -- & -- & -- & 51.72	&56.93	&60.71
 \\
& & & Static Foveated & 45.58& 50.65 & 56.78 & +2.87&+3.85&+9.17&53.20&59.12&66.28 \\
& & & Sunflower Inspired  & 43.23 & 46.28 & 51.61 & -2.44&-5.11&-0.77&50.46&54.02&60.24 \\
& & & Radial Sampling & 45.58 & 47.16 & 48.90 &+2.87&-3.30&-5.98&53.20&55.05&57.08 \\
& & & \textbf{LLMind (Ours)} & \textbf{51.18} & \textbf{61.53} & \textbf{69.74} & \textbf{+15.50} & \textbf{+26.16} & \textbf{+34.09} & \textbf{59.74} & \textbf{71.82} & \textbf{81.41} \\
\cmidrule(lr){2-13}
& \multirow{5}{*}{SmolVLM} & \multirow{5}{*}{78.60}
  & Uniform Sampling               & 39.78 & 45.06 & 48.73 & -- & -- & -- & 50.61&	57.33&	62.00 \\
& & & Static Foveated   & 42.09 & 46.98 & 51.44 & +5.81&+4.26&+5.56&53.55&59.77&65.45 \\
& & & Sunflower Inspired   & 44.27 & 46.20 & 48.64 & +11.29&+2.53&-0.18&56.32&58.78&61.88 \\
& & & Radial Sampling & 41.48 & 43.58 & 47.33 & +4.27&-3.28&-2.87&52.77&55.45&60.22 \\
& & & \textbf{LLMind (Ours)} & \textbf{64.81} & \textbf{72.63} & \textbf{76.81} & \textbf{+62.92} & \textbf{+61.19} & \textbf{+57.62} & \textbf{82.46} & \textbf{92.40} & \textbf{97.72} \\
\cmidrule(lr){2-13}
& \multirow{5}{*}{LLaVA-OneVision} & \multirow{5}{*}{60.26}
  & Uniform Sampling               & 34.80 & 35.71 & 42.33 & -- & -- & -- & 57.75 &	59.26&	70.25 \\
& & & Static Foveated & 34.23 & 35.28 & 37.27 & -1.64&-1.20&-11.95&56.80&58.55&61.85 \\
& & & Sunflower Inspired  & 27.68 & 30.91 & 36.24 & -20.46&-13.44&-14.39&45.93&51.29&60.14 \\
& & & Radial Sampling & 28.29 & 28.03 & 29.78 & -18.71&-21.51&-29.65&46.95&46.52&49.42 \\
& & & \textbf{LLMind (Ours)} & \textbf{43.01} & \textbf{46.09} & \textbf{53.18} & \textbf{+23.59} & \textbf{+29.07} & \textbf{+25.63} & \textbf{71.37} & \textbf{76.49} & \textbf{88.25} \\
\midrule

\multirow{16}{*}{\rotatebox{90}{\textbf{Seed Bench}}}
& \multirow{5}{*}{Qwen2.5-VL} & \multirow{5}{*}{71.25}
  & Uniform Sampling                                & 35.53 & 41.55 & 48.45 & -- & -- & -- & 49.87	&58.32	&68.00 \\
& & & Static Foveated & 34.75 & 42.05 & 49.4 & -2.20&+1.20&+1.96&48.77&59.02&69.33 \\
& & & Sunflower Inspired  & 36.70 & 40.55 & 44.6 & +3.29&-2.41&-7.95&51.51&56.91&62.60 \\
& & & Radial Sampling & 35.20 & 39.45 & 41.75 & -0.93&-5.05&-13.83&49.40&55.37&58.60 \\
& & & \textbf{LLMind (Ours)} & \textbf{40.28} & \textbf{50.50} & \textbf{58.23} & \textbf{+13.37} & \textbf{+21.54} & \textbf{+20.19} & \textbf{56.53} & \textbf{70.88} & \textbf{81.73} \\
\cmidrule(lr){2-13}
& \multirow{5}{*}{SmolVLM} & \multirow{5}{*}{51.6}
  & Uniform Sampling               & 21.81 & 26.38 & 29.19 & -- & -- & -- & 42.27&51.12&56.57 \\
& & & Static Foveated   & 30.20 & 36.85 & 42.65 & +38.47&+39.69&+46.11&58.53&71.41&82.66 \\
& & & Sunflower Inspired   & 33.60 & 34.95 & 37.85 & +54.06&+32.49&+29.67&65.12&67.73&73.35 \\
& & & Radial Sampling & 30.80 & 33.15 & 36.05 & +41.22&+25.66&+23.50&59.69&64.24&69.86 \\
& & &  \textbf{LLMind (Ours)} & \textbf{37.42} & \textbf{43.44} & \textbf{45.97} & \textbf{+71.57}&\textbf{+64.67}&\textbf{+57.49}&\textbf{72.52}&\textbf{84.19}&\textbf{89.09} \\
\cmidrule(lr){2-13}
& \multirow{5}{*}{LLaVA-OneVision} & \multirow{5}{*}{37.75}
  & Uniform Sampling               & 18.65 & 21.60 & 24.35 & -- & -- & -- & 49.40&57.22&64.50 \\
& & & Static Foveated   & 24.85 & 25.55 & 25.60 & +33.24&+18.29&+5.13&65.83&67.68&67.81 \\
& & & Sunflower Inspired   & 21.40 & 23.90 & 25.15 & +14.75&+10.65&+3.29&56.69&63.31&66.62 \\
& & & Radial Sampling & 22.90 & 21.70 & 22.55 & +22.79&+0.46&-7.39&60.66&57.48&59.74 \\
& & &  \textbf{LLMind (Ours)} & \textbf{26.05} & \textbf{27.85} & \textbf{30.40} & \textbf{+39.68}&\textbf{+28.94}&\textbf{+24.85}&\textbf{69.01}&\textbf{73.77}&\textbf{80.53} \\
\bottomrule
\end{tabular}
}
\vspace{-13pt}
\label{tab:scene_level_res}
\end{table*}

\section{Experiments and Evaluation}
\label{sec:experiment}

\subsection{Settings and Implementation Details}
\label{sec:setting_and_implementation}

\noindent\textbf{Evaluation Details.}
\label{sec:eval_details}
We evaluate the LLMind across two complementary VQA paradigms that jointly examine its capability for holistic scene understanding as well as region-guided reasoning. For all experiments, we restrict our study to \textit{lightweight VLMs} ($<$ 4B parameters) to emphasize the framework’s adaptability in low-resource scenarios. This choice reflects a \textit{practical motivation}: recent works show that small-scale models, with efficient design choices and sampling mechanisms, can approach their larger counterparts \cite{gao2024mini, marafioti2025smolvlm}. Our experiments, therefore, highlight how the proposed method enhances reasoning ability without relying on massive model capacity. 

\textbf{(1) Scene-level VQA.} We evaluate the model’s capability for global reasoning over the entire visual scene. The questions demand comprehension of semantic relationships, contextual attributes, and high-level visual concepts present across the image. For example, queries such as \textit{``What best describes the pool of water?''} or \textit{``How many people are visible in the image?''} require a holistic understanding spanning the full spatial extent. We conduct experiments on 5000+ question-answer pairs from \textit{VQAv2} and 2000+ pairs from \textit{Seed-Bench} to evaluate general scene-level reasoning capabilities for LLMind. Additionally, we evaluate LLMind on \textit{A-OKVQA} dataset, which requires integrating external, world-based knowledge for comprehensive visual understanding. 

\textbf{(2) Region-guided VQA.} We evaluate the model’s localized reasoning by using explicit region prompts to direct the model’s focus toward relevant spatial areas. Each question is associated with a bounding region that defines a \textit{cognitive window}. The questions require interpreting spatial cues to perform region-specific classification (RSC) within the defined cognitive window. For example, \textit{``Is the object $\langle$location$\rangle$ a $\langle$class A$\rangle$ or a $\langle$class B$\rangle$?''} or \textit{``Can you provide a description of the region $\langle$location$\rangle$?''}. We adopt data configuration as ControlMLLM~\cite{wu2024controlmllm} for this setting, following the Ferret~\cite{you2023ferret} setup to construct 1400+ region-guided questions from the \textit{LVIS}~\cite{gupta2019lvis} validation split. \vspace{5pt}
\\ 
\textbf{Implementation Details.}
\label{sec:implementation_details}
All experiments were conducted on a system equipped with 4$\times$ NVIDIA GeForce RTX 5090 GPUs (32 GB VRAM each). For each image, we perform an inference-time optimization for a fixed number of iterations $t$. Within this process, the SPSA update step is applied every $t/5$ iterations in equal intervals. Unless otherwise specified, all images are processed at their original resolution without resizing. 

We adopt the concept of \textbf{information-matched images} proposed by Gizdov \textit{et al.}~\cite{gizdov2025seeing}, which provides a principled way to compare sampling strategies under identical pixel budgets. Specifically, two sampling maps are constructed such that they contain the same number of sampled pixels, differing only in how those samples are spatially distributed. This ensures that any observed performance differences arise purely from the sampling distribution rather than from differences in information content. 

The computational overhead for LLMind is marginal and bounded by $t$, with latency of $\sim 0.011$ s/iter w/o SPSA and $\sim 4.75$ s/iter with SPSA (reducible via optimizations~\cite{wu2024controlmllm}).

\begin{figure}[t!]
    \centering
    \includegraphics[width=1.0\linewidth]{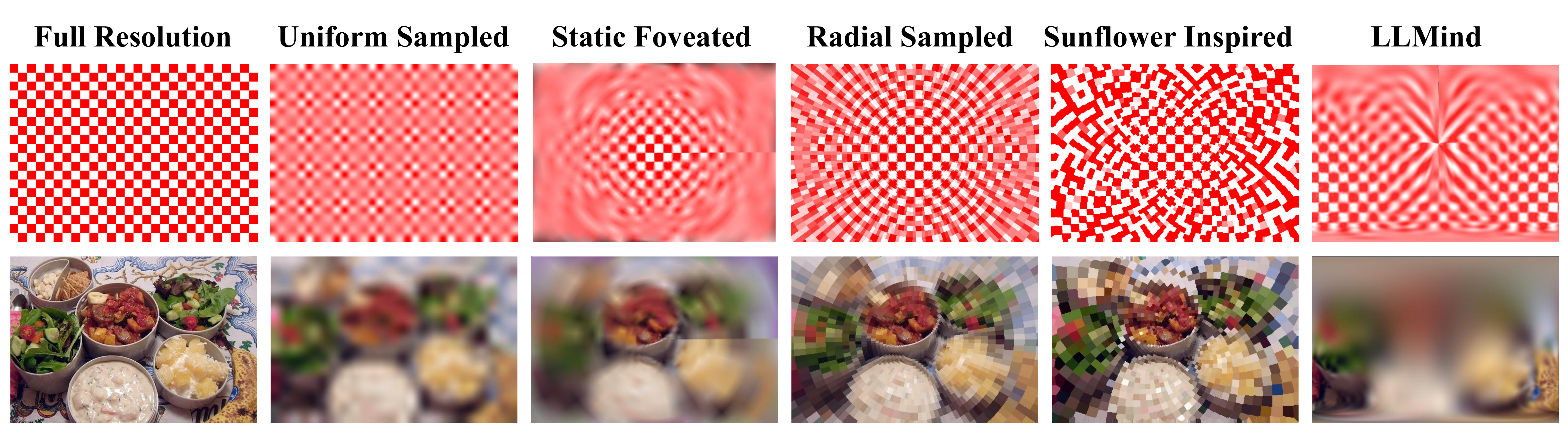}
    \vspace{-18pt}
    \caption{Illustration of the compared sampling methods under 10\% pixel budget. \textit{Zoom in for a better view}.}
    \vspace{-15pt}
    \label{fig:foveation_baselines}
\end{figure}

\subsection{Baseline Methods}
\label{sec:baselines}

Fig.~\ref{fig:foveation_baselines} illustrates the sampling baselines evaluated in our study alongside LLMind. Each method allocates the same total information budget but differs in how resolution is spatially distributed across the visual field. After sampling, all representations are upsampled via interpolation to the original input resolution, ensuring that any performance differences arise purely from the sampling distribution rather than from differences in input size.

\textbf{(1) Uniform Sampling.}
We include a uniform sampling condition as a baseline, where the same total pixel budget is spread evenly across the entire image. 

\textbf{(2) Static Foveated.} We implement the static-foveated baseline following the formulation of Gizdov \etal~\cite{gizdov2025seeing, gizdov2024variable}, using log-polar sampling to generate variable-resolution inputs. Consistent with the original setting, the fixation point is fixed at the image center throughout all experiments.

\textbf{(3) Sunflower Inspired.}
We adopt a biologically inspired sampling layout following the sunflower phyllotaxis model proposed by Killick \etal~\cite{killick2023foveation}. Unlike the original work, we employ this sampling formulation purely as a static input transformation without the graph-convolutional processing. Following the static foveate setting, the fixation is fixed at the image center.

\textbf{(4) Radial Sampling.}
We implement a static radial sampling scheme inspired by the retinal model of Robert-Inacio \etal~\cite{robert2010biologically}. The fixation is fixed at the image center, and the foveal region is preserved at full resolution, while maintaining the overall information-matched constraint so that the total number of sampled pixels remains consistent.

\begin{table*}[t]
\centering
\begin{minipage}{0.48\linewidth}
    \centering
    \caption{Performance comparison of various sampling methods for a single RSC-based VQA task on the LVIS dataset.}
    \vspace{-7pt}
    \resizebox{\linewidth}{!}{
    \begin{tabular}{lclcccc}
    \toprule
    \multirow{3}{*}{\textbf{Model}} & \multirow{2}{*}{\textbf{Accuracy (\%)}} & \multirow{2}{*}{\textbf{Sampling}} &
    \multicolumn{4}{c}{\textbf{Accuracy Sampled}} \\
    & \multirow{2.5}{*}{\textbf{Full Res.}}&  \multirow{2.5}{*}{\textbf{Method}} & \multicolumn{4}{c}{\textbf{(\%)}}\\
    \cmidrule(lr){4-7}
    & & & \textbf{1\%} & \textbf{3\%} & \textbf{5\%} & \textbf{10\%} \\
    \midrule
    
    \multirow{5}{*}{Qwen3-VL} & \multirow{5}{*}{50.60} & Uniform Sampling & 44.80 & 45.63 & 50.40 & 51.60 \\
    & & Static Foveated & 46.10 & 48.50 & 49.40 & 50.40 \\
    & & Sunflower Inspired & 47.60 & 47.50 & 49.70 & 49.50 \\
    & & Radial Sampling & 45.30 & 46.70 & 48.10 & 48.20 \\
     & &  \textbf{LLMind (Ours)} & \textbf{54.60} & \textbf{62.90} & \textbf{71.40} & \textbf{77.50} \\
    \midrule
    \multirow{5}{*}{DeepSeek-VL} & \multirow{5}{*}{46.10} & Uniform Sampling & 43.50 & 43.90 & 44.10 & 47.80\\
    & & Static Foveated & 44.50 & 45.00 & 44.80 & 45.20 \\
    & & Sunflower Inspired & 43.90 & 44.40 & 44.60 & 44.50 \\
    & & Radial Sampling & 44.30 & 44.30 & 45.20 & 44.43 \\
     & &  \textbf{LLMind (Ours)} & \textbf{53.20} & \textbf{59.20} & \textbf{62.20} & \textbf{66.80}  \\
    
    \bottomrule
    \end{tabular}
    }
    \vspace{-12pt}
    \label{tab:roc_res}
\end{minipage}
\hfill
\begin{minipage}{0.48\linewidth}
    \centering
    \caption{Performance comparison of various sampling methods for multiple RSC-based VQA tasks on the LVIS dataset.}
    \vspace{-7pt}
    \resizebox{\linewidth}{!}
    {
    \begin{tabular}{lclcccc}
    \toprule
    \multirow{3}{*}{\textbf{Model}} & \multirow{2}{*}{\textbf{Accuracy (\%)}} & \multirow{2}{*}{\textbf{Sampling}} &
    \multicolumn{4}{c}{\textbf{Accuracy Sampled}} \\
    & \multirow{2.5}{*}{\textbf{Full Res.}}&  \multirow{2.5}{*}{\textbf{Method}} & \multicolumn{4}{c}{\textbf{(\%)}}\\
    \cmidrule(lr){4-7}
    & & & \textbf{1\%} & \textbf{3\%} & \textbf{5\%} & \textbf{10\%}  \\
    \midrule
    
    \multirow{5}{*}{Qwen3-VL} & \multirow{5}{*}{60.20} & Uniform Sampling & 50.90 & 53.49 & 55.30 & 56.85 \\
    & & Static Foveated & 50.38 & 52.19 & 53.74 & 56.58 \\
    & & Sunflower Inspired & 51.16 & 53.22 & 55.29 & 56.07 \\
    & & Radial Sampling & 48.83 & 50.64 & 51.42 & 56.07 \\
     & &  \textbf{LLMind (Ours)} & \textbf{60.21} & \textbf{66.41} & \textbf{72.61} & \textbf{77.26} \\
    \midrule
    \multirow{5}{*}{DeepSeek-VL} & \multirow{5}{*}{51.93} & Uniform Sampling & 43.15 & 45.22 & 47.29 & 50.39 \\
    & & Static Foveated & 44.96 & 48.57 & 52.45 & 51.42 \\
    & & Sunflower Inspired & 42.63 & 47.02 & 48.57 & 49.87 \\
    & & Radial Sampling & 43.15 & 45.21 & 45.73 & 50.38 \\
     & &  \textbf{LLMind (Ours)} &   \textbf{50.90} & \textbf{60.47} & \textbf{65.12} & \textbf{65.63} \\
    
    \bottomrule
    \end{tabular}
    }
    \vspace{-12pt}
    \label{tab:mroc_res}
\end{minipage}
\end{table*}

\begin{figure}[t!]
    \centering
    \includegraphics[width=0.9\linewidth]{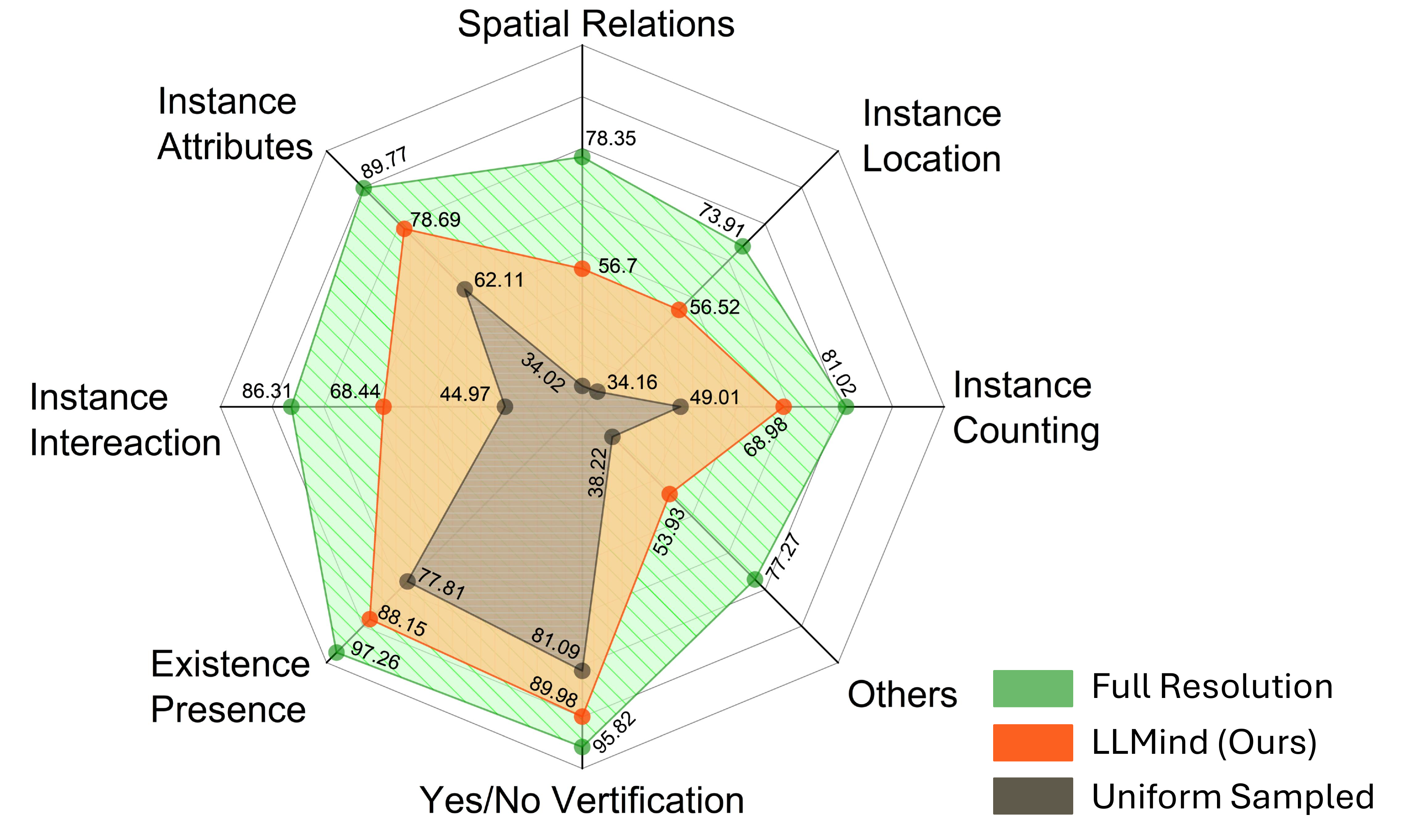}
    \vspace{-8pt}
    \caption{Category-wise performance on A-OKVQA at a 5\% pixel budget with Qwen2.5-VL (4B). \textit{Zoom in for a better view}. } 
    \label{fig:question_type_analysis}
    \vspace{-12pt}
\end{figure}

\subsection{Results and Discussion}
\label{sec:results}

Our experimental results are structured around the following research questions (RQs), which guide our evaluation:

\begin{center}
\vspace{-5pt}
\colorbox{gray!10}{
\begin{minipage}{0.95\linewidth}
\noindent \textbf{RQ-1:} Does LLMind improve scene-level VQA accuracy across diverse benchmarks by selectively allocating pixels to semantically informative regions?

\noindent \textbf{RQ-2:} Can LLMind enhance region-guided VQA accuracy by reducing background distraction and improving evidence grounding?

\noindent \textbf{RQ-3:} What is the impact of the CSF module on sampling quality across different pixel budgets, and how essential is it for semantic alignment?
\end{minipage}
}
\vspace{-5pt}
\end{center}

\subsubsection{Comparison on Scene-level VQA.}
Table~\ref{tab:scene_level_res} presents the quantitative results on VQAv2, Seed-Bench, and A-OKVQA under strict pixel budgets (1--5\%), evaluated across Qwen2.5-VL~\cite{bai2025qwen2} (4B), SmolVLM~\cite{marafioti2025smolvlm} (2B), and LLaVA-OneVision~\cite{li2024llava} (0.9B).

\begin{center}
\vspace{-5pt}
\colorbox{gray!10}{
\begin{minipage}{0.95\linewidth}
\noindent \textcolor{DeepPink}{\textbf{[A-1]}} 
LLMind consistently outperforms all competing sampling strategies under strict pixel budgets. It retains more than \textbf{55\%} of the full-resolution accuracy at only 1\% pixels, over \textbf{70\%} at 3\%, and above \textbf{80\%} at 5\% across all benchmarks. At peak efficiency, LLMind preserves up to \textbf{82\%}, \textbf{92\%}, and \textbf{97\%} of the full-resolution performance at the 1\%, 3\%, and 5\% budgets.
\end{minipage}
}
\vspace{-10pt}
\end{center}

Fig.~\ref{fig:qual_vqa} provides qualitative comparisons on Seed-Bench for Qwen2.5-VL, including attention visualizations. LLMind’s attention closely aligns with the full-resolution model, consistently focusing on the evidence-relevant regions, whereas uniform sampling produces diffuse and often misplaced attention.
Moreover, the question-category-wise analysis (Fig.~\ref{fig:question_type_analysis}) reveals that LLMind improves performance across all major reasoning categories, indicating broad gains that extend beyond any specific task type.

\subsubsection{Comparison on Region-guided VQA.}
Table~\ref{tab:roc_res} and Table~\ref{tab:mroc_res} present results for single- and multi-region guided VQA on the LVIS dataset, where each question specifies a cognitive window. We evaluate Qwen3-VL~\cite{yang2025qwen3} (2B) and DeepSeek-VL~\cite{lu2024deepseek} (2B) in this setting, which requires performing region-specific classification (RSC) through VQA within the cognitive window.

\begin{center}
\vspace{-2pt}
\colorbox{gray!10}{
\begin{minipage}{0.95\linewidth}
\noindent \textcolor{DeepPink}{\textbf{[A-2]}} 
LLMind consistently surpasses all baselines under both single and multi-region guided VQA. With only \textbf{3\%} pixels, LLMind even \textbf{outperforms the full-resolution input}, indicating that removing irrelevant visual content can reduce distraction and improve reasoning fidelity.
\end{minipage}
}
\vspace{-8pt}
\end{center}

The targeted visual allocation in LLMind explains its ability to surpass full-resolution inputs in region-guided tasks. Small VLMs such as Qwen3 (2B) and DeepSeek-VL (2B) achieve below 60\% accuracy at full resolution, indicating limited robustness in this setting. By revealing only evidence-relevant regions and suppressing background clutter,
LLMind reduces the influence of background clutter and scene-level priors that often lead to hallucinations. As a result, the model makes decisions that are more closely grounded in the specified regions, leading to more reliable and accurate reasoning.

\subsubsection{Ablation and Analysis}

\begin{figure}[h!]
    \centering
    \vspace{-5pt}
    \includegraphics[width=0.90\linewidth]{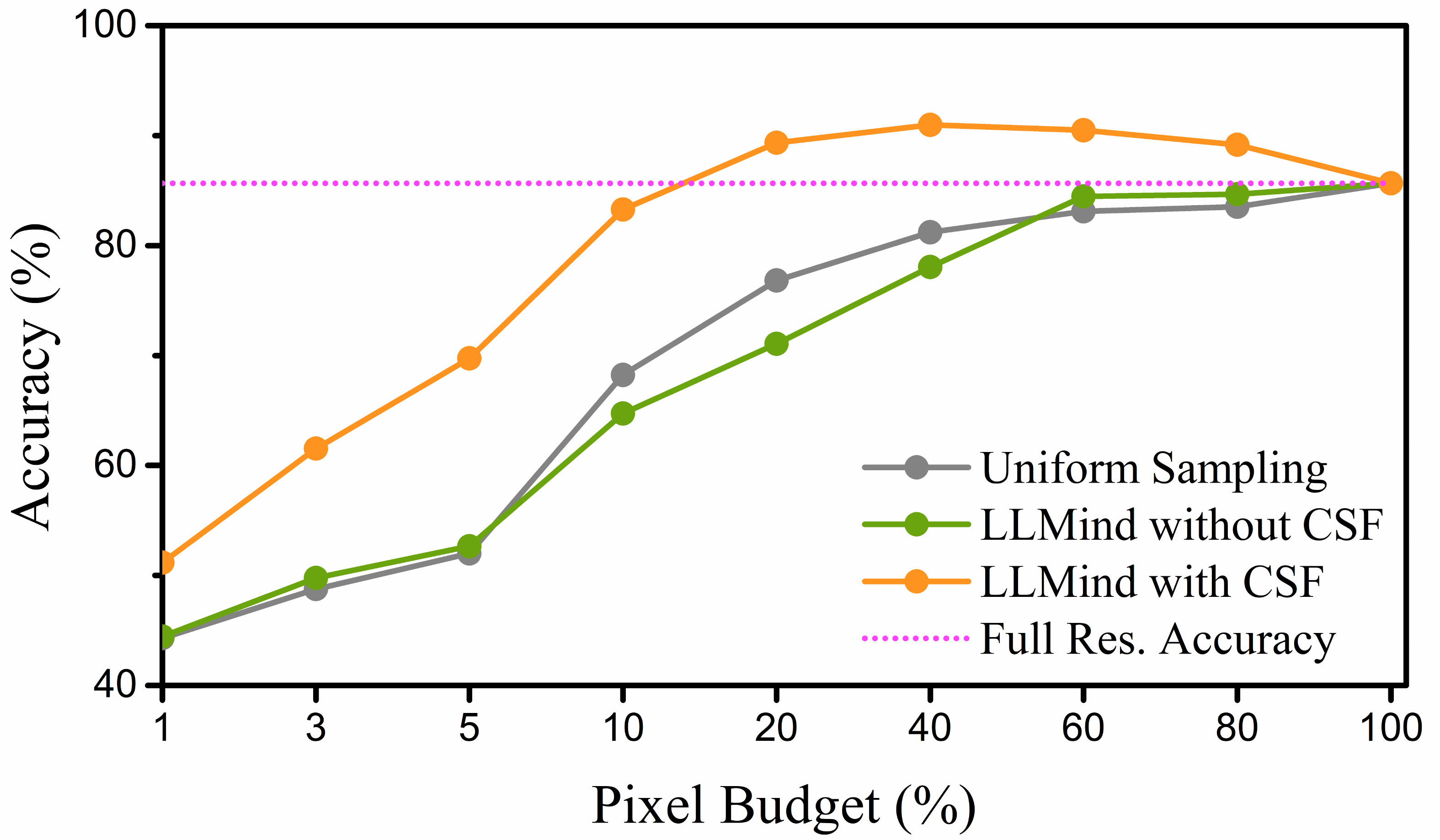}
    \vspace{-5pt}
    \caption{Ablation study on A-OKVQA dataset with Qwen2.5-VL (4B) illustrating the impact of pixel budget and the CSF module on LLMind's performance. \textit{Zoom in for a better view}.}
    \vspace{-12pt}
    \label{fig:ablation}
\end{figure}

We conduct an ablation study on the A-OKVQA dataset using Qwen2.5-VL to analyze the roles of pixel budget and the CSF module (Fig.~\ref{fig:ablation}). 
\begin{center}
\vspace{-5pt}
\colorbox{gray!10}{
\begin{minipage}{0.95\linewidth}
\noindent \textcolor{DeepPink}{\textbf{[A-3]}} 
Without CSF, LLMind relies only on perceptual similarity, yielding gains over Uniform Sampling at low pixel budgets but diminishing at higher budgets where perceptual cues fail to highlight semantic regions. With CSF, performance improves consistently across all pixel budgets, showing that semantic feedback is essential for guiding sampling toward task-relevant evidence.
\end{minipage}
}
\vspace{-8pt}
\end{center}

With CSF, LLMind achieves even stronger gains: at only 20\% of the sampled pixels, it \textbf{surpasses the full-resolution accuracy}, indicating that VLMs do not inherently require the entire image to perform well. When sampling is guided adaptively toward semantically relevant regions, the model can match and even exceed the performance obtained from full-resolution inputs. We also observe \textbf{diminishing returns} beyond the 20\% pixel budget. The accuracy gradually converges toward the full-resolution performance as the pixel budget increases.

\section{Conclusion and Future Work}
\label{sec:conclusion}

In this work, we investigated the role of biologically inspired sampling in enhancing visual reasoning under constrained pixel budgets. We introduced \textbf{LLMind}, a training-free adaptive sampling strategy inspired by cortical magnification in human vision. 
Comprehensive evaluations across diverse datasets and multiple backbone VLMs demonstrate that LLMind consistently outperforms uniform and other bio-inspired baseline, in both \emph{scene-level} and \emph{region-guided} VQA settings, highlighting its utility for efficient and scalable multimodal intelligence. 

\noindent \textbf{Future Work:}
Looking forward, we aim to extend this framework toward more generative and 3D-aware vision-language tasks, including object-centric descriptions, 3D VQA, and compression-aware perception for neural rendering pipelines. We hope this work encourages further exploration of perceptually grounded representations in VLMs.


\clearpage
{
    \section*{Acknowledgments} This work is supported by the MOE AcRF Tier 1: Call2/2025 under Grant No. RG160/25 and NTU Start Up Grant.
    \small
    \bibliographystyle{ieeenat_fullname}
    \bibliography{main}
}


\end{document}